\documentclass[journal,twoside,web]{ieeecolor}
\usepackage{generic}

\usepackage{amsmath,amssymb,amsfonts}
\usepackage{algorithmic}
\usepackage{graphicx}
\usepackage{algorithm,algorithmic}
\usepackage{textcomp}
\usepackage{array}
\usepackage[backend=biber, style=authoryear]{biblatex}

\def\BibTeX{{\rm B\kern-.05em{\sc i\kern-.025em b}\kern-.08em
    T\kern-.1667em\lower.7ex\hbox{E}\kern-.125emX}}
\begin{document}
\title{OpenECG: Benchmarking ECG Foundation Models with Public 1.2 Million Records}
\author{Zhijiang Wan\textsuperscript{1,2,†}, Qianhao Yu\textsuperscript{2,†}, Jia Mao, Wenfeng Duan\textsuperscript{1,*},  Cheng Ding\textsuperscript{3,*}
\thanks{This work was supported in part by the National Natural Science
Foundation of China under Grant 62461039, in part by the Jiangxi
Provincial Natural Science Foundation under Grant 20242BAB25059
and 20232BAB212029. (Zhijiang Wan and Qianhao Yu are co-first authors.) (Corresponding author:
Wenfeng Duan and Cheng Ding)}
\thanks{Zhijiang Wan and Wenfeng Duan are with the Department of Radiology, The First Affiliated Hospital, Jiangxi Medical College, Nanchang University, Nanchang 330006, China (email: zhijiangwan@ncu.edu.cn; ndyfy02345@ncu.edu.cn).}
\thanks{ Qianhao Yu and Jia Mao are with the School of Information Engineering, Nanchang University, Jiangxi, China (e-mail: 3207270132@qq.com; 1782513101@qq.com).}
\thanks{Cheng Ding is with Department of Biomedical Engineering, Georgia Institution of Technology, Atlanta 30332-0315, United States (e-mail: chengding@gatech.edu).}
}

\maketitle

\begin{abstract}
This study introduces OpenECG, a large-scale benchmark of 1.2 million 12-lead ECG recordings from nine centers, to evaluate ECG foundation models (ECG-FMs) trained on public datasets. We investigate three self-supervised learning methods (SimCLR, BYOL, MAE) with ResNet-50 and Vision Transformer architectures, assessing model generalization through leave-one-dataset-out experiments and data scaling analysis. Results show that pre-training on diverse datasets significantly improves generalization, with BYOL and MAE outperforming SimCLR, highlighting the efficacy of feature-consistency and generative learning over contrastive approaches. Data scaling experiments reveal that performance saturates at ~60-70\% of total data for BYOL and MAE, while SimCLR requires more data. These findings demonstrate that publicly available ECG data can match or surpass proprietary datasets in training robust ECG-FMs, paving the way for scalable, clinically meaningful AI-driven ECG analysis.
\end{abstract}


\section{Introduction}
\label{sec:introduction}
Electrocardiography (ECG) is a fundamental tool for diagnosing cardiovascular diseases (CVDs), which are among the leading causes of mortality worldwide. ECG enables clinicians to detect arrhythmias, myocardial infarction, and other heart conditions\parencite{bib1}. Despite its importance, several challenges hinder the effective utilization of ECG in clinical practice: First, the diagnostic accuracy can differ significantly among cardiologists due to varying levels of training and experience. Second, continuous ECG monitoring generates vast amounts of data, making it difficult for cardiologists to analyze and interpret manually within a reasonable timeframe. Artificial intelligence (AI) offers promising solutions to these challenges. Recent advancements in AI, particularly deep learning, have demonstrated remarkable potential in automating ECG analysis, improving diagnostic accuracy, and reducing the burden on clinicians \parencite{bib2}.

Among these advancements, the concept of foundation models (FMs) \parencite{bib3} has emerged as a transformative approach. Foundation models, which are large-scale pre-trained models, have revolutionized various domains, including natural language processing and computer vision. Examples such as GPT \parencite{bib4} for text and CLIP \parencite{bib5} for vision demonstrate how pre-training on diverse datasets can enable models to generalize across multiple tasks. Inspired by this success, several studies have explored the application of foundation models for ECG analysis.

Apple developed the first ECG foundation model using ECG data from 106,643 participants \parencite{bib6}. However, this model is limited to single-lead I ECG and was tested on the same cohort. Additionally, its primary tasks—predicting age, BMI, and sex—do not include heart rhythm analysis. Other foundation models, such as AnyECG \parencite{bib7} and ECG-FM \parencite{bib8}, combined data from multiple sources but failed to evaluate their generalizability effectively. Beyond the conventional self-supervised approach to training ECG foundation models, another strategy involves incorporating corresponding diagnostic reports. Models like ECG-Chat \parencite{bib9}, ECG Semantic Integrator \parencite{bib10}, and MERL \parencite{bib11} leverage ECG-text pairs from MIMIC-IV-ECG \parencite{bib12}. However, since these datasets are collected from a single center, their ability to generalize across diverse populations and recording environments is limited. This underscores the need for multi-center, diverse datasets to develop more robust and widely applicable ECG foundation models.

To address this limitation, our study systematically curated all publicly available 12-lead ECG datasets, integrating data from multiple institutions worldwide. These datasets encompass 1,233,337 ECG recordings from 483,837 patients across 9 centers, including both annotated clinical diagnoses and unannotated raw signals for self-supervised learning. By aggregating data from diverse sources, we aim to build an ECG foundation model that is not only technically superior but also clinically meaningful—one that can assist physicians across different healthcare settings with reliable and unbiased diagnostic support.

A key challenge in developing clinically useful AI models is determining the most effective learning paradigm for ECG analysis. In this study, we evaluate three distinct self-supervised learning approaches to building ECG foundation models: (1) SimCLR \parencite{bib13}, which learns representations by distinguishing between augmented versions of the same ECG signal; (2) BYOL (Bootstrap Your Own Latent) \parencite{bib14}, which removes negative sample constraints and focuses on feature consistency across transformations; and (3) Masked Autoencoder (MAE) \parencite{bib15}, a generative approach that reconstructs missing ECG segments, simulating real-world noisy or incomplete recordings. By comparing these methods, we aim to identify the most clinically relevant approach for learning robust ECG representations that generalize across institutions and patient populations.

Instead of finding a ‘winner model’, this study represents the first comprehensive effort to unify all publicly available 12-lead ECG datasets into a single benchmark for foundation model training. More importantly, it underscores the vast yet underutilized potential of open datasets in AI-driven ECG research. Contrary to the prevailing notion that high-performing AI models require proprietary data, we demonstrate that publicly available data, when curated and leveraged effectively, can yield clinically valuable AI models capable of supporting cardiologists in their daily practice. 

\begin{figure*}[!t]  
    \centering
    \includegraphics[scale=0.7]{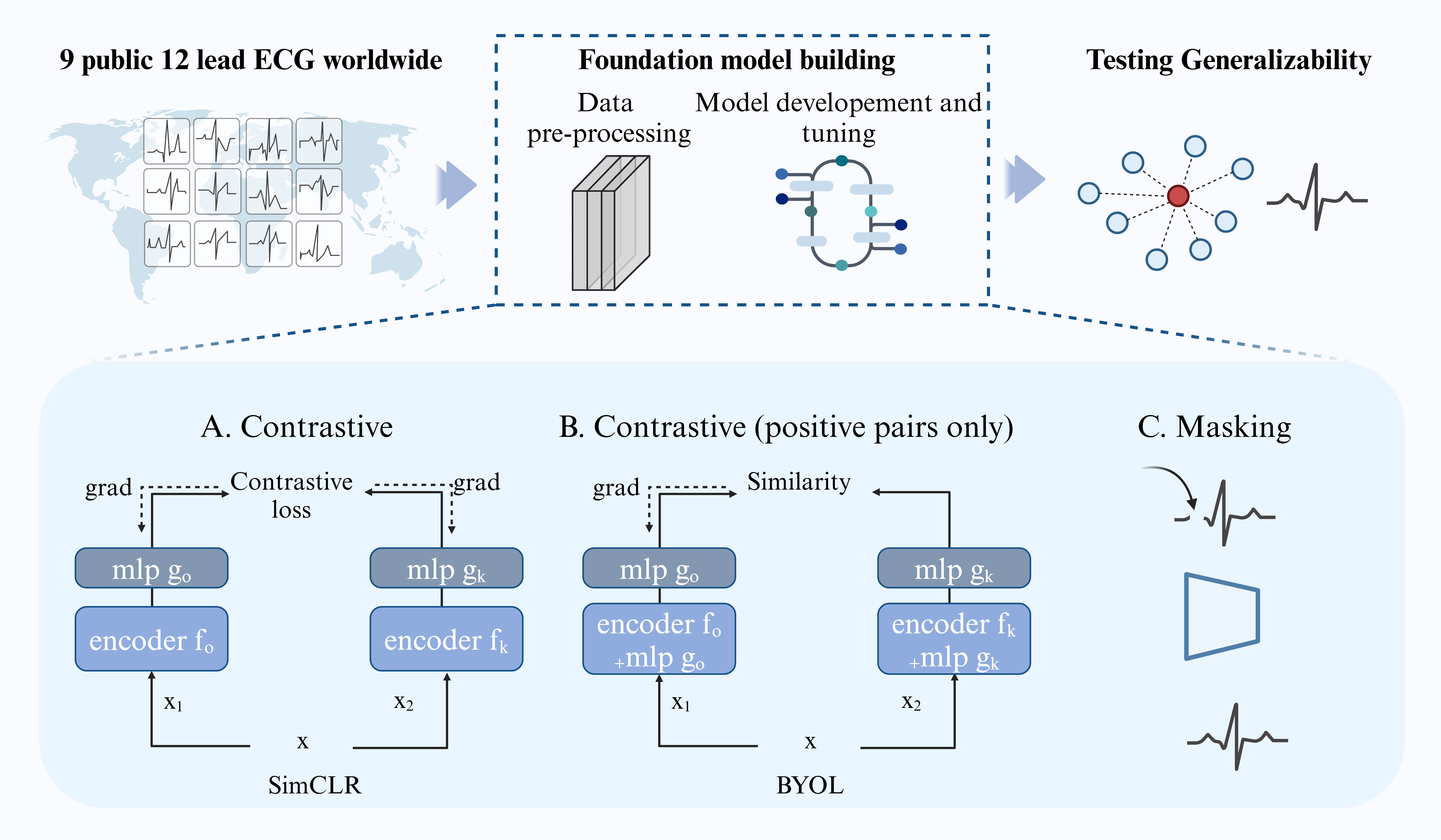}  
    \caption{The key dataset statistics of previous ECG Foundation Models.}  
    \label{fig:example 1}  
\end{figure*}

\section{Related work}
\subsection{Architecture design of FMs for ECGs}
Transformer-based and CNN-based backbones are the two primary approaches for constructing pre-trained foundation models (FMs) in ECG signal processing. Our previous work \parencite{bib16} categorizes the current Transformer-based backbones for ECG-FMs into seven distinct groups: ViT-series (Vision Transformer), Bidirectional Encoder Representations from Transformers (BERT-series), Contrastive Language–Image Pre-training (CLIP-series), Generative Pre-trained Transformer (GPT-series), Large Language Model Meta AI (LLaMA-series), and DALL-E-series. Prior to the widespread adoption of Transformers, various traditional pre-training techniques employed deep neural networks (e.g., CNN, RNN, and MLP) and their variants as the basic architectures for pre-training FMs. Given that ECG signals are time-series data, 1D convolutional layers and pooling operations are commonly employed to capture short-term temporal patterns, such as those related to heart rhythm. Pooling operations, in particular, play a crucial role in reducing the dimensionality of the data while preserving key features, such as the R-peak information. 

Unlike previous studies that focus on designing advanced and specialized architectures tailored for processing ECGs, our approach diverges by choosing commonly used FM architectures. We aim to investigate the effect of combining various pre-trained models, each built on different architecture types and learning strategies, and applying them across multiple-center ECG datasets. This approach allows us to explore the generalization capability of these FMs, particularly in terms of how well they adapt to diverse datasets and real-world variability. By evaluating the impact of different pre-training strategies on model performance across multiple centers, we seek to shed light on the robustness of these models and their potential for more generalized ECG signal processing tasks.
\subsection{SSL-based pre-training strategies of ECG-FMs}
Unlike traditional supervised learning, which relies heavily on labeled datasets that can be costly and time-consuming to curate, SSL leverages the inherent structure of the data itself to generate meaningful training signals. This approach reduces the dependence on manual annotations and allows for the utilization of large-scale, real-world datasets that are often unlabeled. In NLP field, SSL techniques such as masked language modeling in BERT or autoregressive learning in GPT have demonstrated remarkable success in capturing semantic and syntactic nuances. 

In computer vision field, contrastive learning methods like SimCLR \parencite{bib13} and MoCo \parencite{bib17} enable models to learn rich visual representations without labeled images. Similarly, in biomedical fields, SSL facilitates the pre-training of large models on data such as medical images or physiological signals, where labeled datasets are scarce or difficult to annotate. Notably, SSL techniques like BERT \parencite{bib18} and GPT from NLP and SimCLR and MoCo from CV can be adapted and transferred to the domain of ECG processing. For instance, Song et al. \parencite{bib19} concluded that contrastive learning and generative learning are two fundamental SSL approaches for pre-training FMs. They developed a hybrid SSL method that effectively combines both paradigms to enhance the pre-training of their FMs, leveraging the strengths of each approach for improved representation learning.  McKeen et al. \parencite{bib8} integrated the masking objective technique with contrastive learning to pre-train their ECG-FMs. Based on previous work, we aim to investigate the impact of SSL-based pre-training strategies combined with existing FM architectures on the generalization ability of FMs. 

To this end, we selected three representative SSL methods (i.e., SimCLR, BYOL, and MAE) as pre-training strategies. SimCLR, which leverages positive and negative sample pairs, excels in extracting meaningful and fine-grained features from high-dimensional data in an unsupervised manner. In contrast, BYOL eliminates the need for negative samples, focusing solely on optimizing the model's feature representation through self-supervised objectives, thereby simplifying training and improving robustness. Meanwhile, MAE, as a masking-based autoencoder technique, trains the model by randomly masking parts of the input and requiring it to reconstruct the original data. This encourages the model to capture both the local and global structures of the data. 

\section{Materials and Methods}

\subsection{Data preparation}
\subsubsection{Dataset description}

\begin{table*}[ht]
\caption{The data details in the previous ECG foundation model studies}
\centering
\begin{tabular}{p{2cm}ccccccc}
\hline
\textbf{ } & \textbf{\#Participants} & \textbf{\#ECG recordings}&  \textbf{\#Centers for training data} & \textbf{\#Centers for testing data}  & \textbf{\#ECG lead}  & \textbf{\#Classes}  \\
\hline
Apple Heart and Movement Study \parencite{bib6} & 106,643 & 3,743,679 & 1 & 1 & 1 & 2 \\

AnyECG \parencite{bib7} & 53,563 & 53,101 & 6 & 6 & 12 & 5 \\

ECG-FM \parencite{bib8} & 372,851 & 1,560,494 & 7 & 7 & 12 & 13 \\

ECG-Chat \parencite{bib9} & 161,352 & 800,035 & 1 & 2 & 12 & 16 \\

ECG Semantic Integrator \parencite{bib10} & 161,352 & 800,035 & 3 & 2 & 12 & 5 \\

MERL \parencite{bib11} & 161,352 & 800,035 & 1 & 3 & 12 & 5 \\

ECGFounder \parencite{bib20} & 1,818,247 & 10,771,552 & 1 & 3 & 12 & 150 \\

Ours & 483,837 & 1,233,337 & 8 & 6 & 12 & 24 \\
\hline

\end{tabular}
\end{table*}

Table 1 illustrates the data details in the previous ECG foundation model studies. 
Building upon this foundation,this study integrates multiple publicly available 12-lead ECG datasets to form a comprehensive benchmark dataset to support cardiovascular disease automatic diagnosis research.  Below are the detailed descriptions of the datasets used: 

\textbf{MIMIC-IV-ECG \parencite{bib12}}: The MIMIC-IV dataset, sourced from the MIMIC-IV clinical database, contains approximately 800,000 ECG records from nearly 160,000 patients. All records are 12-lead, sampled at 500Hz, with a duration of 10 seconds, covering the period from 2008 to 2019. The MIMIC-IV-ECG dataset is strongly correlated with other clinical database information (such as demographics, diagnoses, medications, and lab results) and can be used for studying cardiovascular disease diagnosis in emergency departments, wards, and intensive care units (ICU), such as myocardial ischemia, heart attacks, and arrhythmias.

\textbf{CODE-15 \cite{bib21}}: The CODE-15 dataset is a stratified subset of the CODE dataset, containing 345,779 12-lead ECG records from 233,770 patients, spanning from 2010 to 2016. It was collected by the Telehealth Network of Minas Gerais (TNMG) in Brazil and is widely used in ECG automatic diagnosis research. For example, related studies have used deep neural networks for automatic ECG diagnosis and cardiovascular event risk prediction (such as estimating "ECG age" to assess mortality). The scale and annotation quality of CODE-15 provide a solid foundation for ECG AI algorithms.

\textbf{PhysioNet 2020 \parencite{bib22}}: This dataset integrates 12-lead ECG data from multiple sources, including the CPSC database, PTB database, St Petersburg INCART database, Georgia 12-lead database, PTBXL database, and other unpublished databases, covering various cardiovascular conditions and signal features. The sampling rate is mainly 500Hz, with some reaching 1000Hz. The recording durations range from 6 seconds to several minutes. The Challenge 2020 dataset provides rich and diverse data support for cardiovascular disease automatic detection algorithms.

\textbf{Chapman \parencite{bib23}}: The Chapman-Shaoxing dataset was created by Chapman University in collaboration with Shaoxing People's Hospital and Ningbo First Hospital. It includes 12-lead ECG records from 45,152 patients, sampled at 500Hz with a duration of 10 seconds. The dataset underwent two rounds of annotation by certified physicians, and all diagnostic labels (such as atrial fibrillation, premature beats, left bundle branch block, right bundle branch block, etc.) were confirmed by experienced doctors, providing high-quality annotated data for research on automatic classification and prediction models for cardiovascular diseases. 

Table 2 summarizes key characteristics of labeled and unlabeled ECG datasets used in the study, including patient counts, sample sizes, leads, duration (in seconds), and sampling rates (Hz). Labeled datasets include CPSC, Georgia, PTB, PTB-XL, St. Petersburg, and Chapman, while unlabeled datasets include MIMIC and CODE15. These datasets vary in scale, with MIMIC and CODE15 providing large sample counts. Labeled datasets like PTB offer higher resolution and sampling rates, while St. Petersburg has lower sampling rates. This highlights the diversity of data sources used for training and evaluation.

\begin{table*}[ht]
\caption{Key characteristics of labeled and unlabeled ECG datasets used in the study.}
\centering
\begin{tabular}{c|cc|cccccc}
\hline
\multicolumn{1}{c|}{\textbf{}} & \multicolumn{2}{c|}{\textbf{Unlabeled}} & \multicolumn{6}{c}{\textbf{Labeled}} \\

\multicolumn{1}{c|}{\textbf{Demographic Information}} & \textbf{MIMIC} & \multicolumn{1}{c|}{\textbf{CODE15}}&  \textbf{CPSC} & \textbf{Geogria}  & \textbf{PTB}  & \textbf{PTB-XL}  & \textbf{St. Petersb.}  & \textbf{Chapman} \\
\hline
\#Patients & 161,352 & 233,770 & 14,053 & 10,344 & 249 & 18,885 & 32 & 45,152\\

\#Sample & 800,035 & 341,292 & 14,053 & 10,344 & 549 & 21,837 & 75 & 45,152\\

\#Leads & 12 & 12 & 12 & 12 & 12 & 12 & 12 & 12 \\

Length & 5,000 & - & - & 5,000 & - & 5,000 & 462,600 & 5,000\\

Duration(s) & 10 & 7 to 10 & 6 to 60 & 10 & - & 10 & 1,800 & 10\\

Sampling rate(Hz) & 500 & 400 & 500 & 500 & 1,000 & 500 & 257 & 500\\
\hline
\end{tabular}
\end{table*}

\subsubsection{Data pre-processing}
In this study, we preprocess multiple ECG signal datasets to form a unified experimental framework. The Challenge dataset of PhysioNet 2020 is split into five sub-datasets: CPSC, Georgia, PTB, PTB-XL, and St. Peter, to evaluate model performance across different scenarios. We use the MIMIC-IV and CODE-15 datasets as unlabeled pre-training data, while the Challenge and Chapman datasets are used for supervised pre-training and fine-tuning. To avoid the high time cost of real-time data augmentation, offline augmentation is first performed, and lightweight online data augmentation strategies, such as random zeroing, are used to maintain data diversity during training. The data processing strictly retains the original 12-lead structure, with each sample resampled to a fixed length of 1000 points, and the data format unified as ((n, 12, 1000)), where (n) is the number of samples. Additionally, to avoid having the same patient's different samples appear in both the training and validation/test sets, the data is split by patient to ensure dataset independence. Each sub-dataset is divided into five folds for cross-validation experiments. The label system is based on the SNOMED CT standard \parencite{bib24}, selecting 24 ECG-related diseases as classification targets.

\subsection{Pre-training of ECG-FMs}

For the pre-training of ECG foundation models, we employed a five-fold cross-validation strategy to ensure robust model evaluation and avoid overfitting. The entire dataset was divided into five equally sized folds, with each fold used once as a test set while the remaining four folds served as the training set. Within each training set, one fold was iteratively selected as a validation set for monitoring training performance. This approach provides a balanced evaluation across multiple subsets of the data, facilitating the assessment of generalization ability across diverse patient populations and environments. 

To further enhance the diversity of the training data, we employed a masking technique that included both temporal masking and lead masking:

\textbf{Temporal Masking.} This technique randomly masks 100 consecutive data points within the ECG signal, simulating signal interruptions, such as noise or data loss. The temporal mask was applied uniformly across all 12 leads, ensuring consistent removal of data segments across the entire ECG signal.

\textbf{Lead Masking.} This method introduces a random masking process across all 12 leads simultaneously. A segment of 100 data points within each lead was masked, simulating conditions such as multi-lead signal corruption or synchronized disruptions in the recording.

These masking strategies simulate real-world scenarios of incomplete or noisy ECG data, encouraging the models to learn robust representations that can handle missing or corrupted data effectively.

\subsection{Self-Supervised Learning Methods}

We evaluated three state-of-the-art SSL methods to pre-train the ECG foundation models: BYOL, SimCLR and MAE. Each method was paired with two distinct deep learning backbones: ResNet-50 \parencite{bib25} and Vision Transformer (ViT) \parencite{bib26}. BYOL and SimCLR with ResNet-50 Backbone: Both BYOL and SimCLR were used to pre-train the ResNet-50 backbone. These contrastive learning methods are designed to encourage the model to learn invariant representations of the ECG signals by comparing augmented versions of the data, maximizing similarity between similar signals while minimizing the distance between dissimilar ones. MAE with ViT Backbone: The MAE method was adopted to pre-train the ViT backbone. Unlike contrastive learning, MAE focuses on learning to reconstruct masked portions of the ECG signal, simulating real-world scenarios of incomplete or corrupted data. 

\begin{figure*}[!t]  
    \centering
    \includegraphics[scale=0.7]{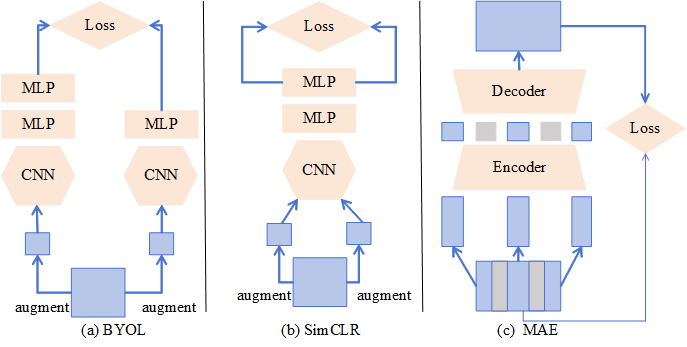}  
    \caption{Three SSL pre-train methods.}  
    \label{fig:example 1}  
\end{figure*}

\section{Experiments and Results}
\subsection{Comparison of ECG rhythm classification performance with previous studies}

\begin{table*}[ht]
\caption{Comparison of ECG rhythm classification performance with previous studies.}
\centering
\begin{tabular}{ccccccc}
\hline
\multicolumn{7}{c}{\textbf{PhysioNet challenge 2020 dataset}}  \\

\multicolumn{1}{c|}{\textbf{}} & \textbf{PTB-XL} & \textbf{CPSC 2018}&  \textbf{Chapman} &  \textbf{PTB}  & \textbf{Geogria}  & \textbf{INCART} \\
\multicolumn{1}{c|}{\textbf{}} & \multicolumn{6}{c}{\textbf{F1 Score/ AUROC}}\\
\hline
\multicolumn{1}{c|}{\textbf{ECG-Chat}} & \textbf{55.9/94.1} & \textbf{80.1 / 95.7}&  \textbf{-} &  \textbf{-}  & \textbf{-}  & \textbf{-} \\
\multicolumn{1}{c|}{\textbf{MERL}} & \textbf{48.1  /91.9} & \textbf{72.8 / 92.6}&  \textbf{- / 87.9} &  \textbf{-}  & \textbf{-}  & \textbf{-} \\
\multicolumn{1}{c|}{\textbf{ESI}} & \textbf{- / 93.1} & \textbf{-}&  \textbf{-} &  \textbf{-}  & \textbf{-}  & \textbf{-} \\
\multicolumn{1}{c|}{\textbf{MAEFE}} & \textbf{64.7 / 88.6} & \textbf{71.6 / 94.5}&  \textbf{-} &  \textbf{-}  & \textbf{-}  & \textbf{-} \\
\multicolumn{1}{c|}{\textbf{Ours-SimCLR}} & \textbf{46.9 / 91.5} & \textbf{73.1 / 92.4}&  \textbf{52.3 / 95.1} &  \textbf{37.8 / 74.2}  & \textbf{12.3 / 72.0}  & \textbf{17.1 / 69.8} \\
\multicolumn{1}{c|}{\textbf{Ours-BYOL}} & \textbf{47.7 / 91.1} & \textbf{72.8 / 92.6}&  \textbf{51.5 / 94.8} &  \textbf{36.1 / 73.4}  & \textbf{26.2 / 68.5}  & \textbf{11.7 / 70.3} \\
\multicolumn{1}{c|}{\textbf{Ours-MAE}} & \textbf{48.1 / 90.9} & \textbf{74.5 / 93.2}&  \textbf{50.8 / 94.2} &  \textbf{35.4 / 72.6}  & \textbf{25.3 / 67.9}  & \textbf{10.5 / 71.2} \\
\hline
\multicolumn{1}{c|}{\textbf{3KG}} & \multicolumn{6}{c}{\textbf{43.2/88.3}}\\
\multicolumn{1}{c|}{\textbf{Patient Contrastive Learning}} & \multicolumn{6}{c}{\textbf{36.2/85.7}}\\
\multicolumn{1}{c|}{\textbf{AnyECG}} & \multicolumn{6}{c}{\textbf{28.3/-}}\\
\multicolumn{1}{c|}{\textbf{ECG-FM}} & \multicolumn{6}{c}{\textbf{22.8/-}}\\
\multicolumn{1}{c|}{\textbf{Ours-SimCLR}} & \multicolumn{6}{c}{\textbf{41.5/89.6}}\\
\multicolumn{1}{c|}{\textbf{Ours-BYOL}} & \multicolumn{6}{c}{\textbf{42.8/87.6}}\\
\multicolumn{1}{c|}{\textbf{Ours-MAE}} & \multicolumn{6}{c}{\textbf{44.1/89.2}}\\
\hline
\end{tabular}
\end{table*}

Table 2 presents a comparison of our model against various other methods on multiple datasets used in the ECG classification task. We evaluate the performance of different models on the PhysioNet Challenge 2020 dataset, as well as the PTB-XL, CPSC 2018, Chapman, PTB, Georgia, and INCART datasets, using metrics such as F1 score and AUROC. Our models, including Ours-SimCLR, Ours-BYOL, and Ours-MAE, consistently show competitive performance across different datasets. For instance, Ours-SimCLR achieves F1 scores and AUROCs of 46.9/91.5, 73.1/92.4, and 52.3/95.1 on the PTB-XL, CPSC 2018, and Chapman datasets, respectively. Meanwhile, Ours-BYOL and Ours-MAE demonstrate similar strong performance across the same datasets. In comparison to other models such as 3KG and Patient Contrastive Learning, our methods yield higher F1 scores and AUROCs, showcasing their effectiveness in ECG signal classification. Overall, our models demonstrate solid generalization across different datasets, reflecting their robustness and adaptability.

\subsection{Testing Generalizability via Iterative Dataset Exclusion}

To assess the model’s ability to generalize across diverse datasets, we employ an iterative leave-one-dataset-out (LODO) strategy. Specifically, we train the ECG foundation model on all available datasets except one, which is then used as an independent downstream test set. This process is repeated for each dataset, ensuring that every dataset serves as a test set once. Performance metrics, including classification accuracy, F1-score, and domain shift analysis, are recorded to quantify the model’s robustness across different data distributions.

\begin{table*}[ht]
\caption{Performance comparison via Iterative Dataset Exclusion from the pre-training dataset.}
\centering
\begin{tabular}{cccccccccc}
\hline
\multicolumn{1}{c|}{\textbf{Target dataset}} & \multicolumn{3}{c|}{\textbf{BYOL}} &  \multicolumn{3}{c|}{\textbf{SimCLR}}&   \multicolumn{3}{c}{\textbf{MAE}} \\
 \multicolumn{1}{c|}{\textbf{}} & \textbf{w/} &  \textbf{w/o} &  \multicolumn{1}{c|}{\textbf{$\triangle$ }} &  \textbf{w/} &  \textbf{w/o} &  \multicolumn{1}{c|}{\textbf{$\triangle$ }}  & \textbf{w/} &  \textbf{w/o} &  \textbf{$\triangle$ }  \\
\hline
\textbf{CPSC} & \textbf{71.9} & \textbf{69.3} & \textbf{3.6\%} & \textbf{72.1} & \textbf{70.2} & \textbf{2.6\%} & \textbf{70.5} & \textbf{68.3} & \textbf{3.1\%} \\
\textbf{PTB-XL} & \textbf{91.1} & \textbf{90.4} & \textbf{0.8\%} & \textbf{90.9} & \textbf{90.1} & \textbf{0.9\%} & \textbf{92.0} & \textbf{91.3} & \textbf{0.8\%} \\
\textbf{Chapman} & \textbf{94.8} & \textbf{94.6} & \textbf{0.2\%} & \textbf{94.5} & \textbf{94.3} & \textbf{0.2\%} & \textbf{94.2} & \textbf{94.1} & \textbf{0.1\%} \\
\textbf{PTB} & \textbf{73.4} & \textbf{72.6} & \textbf{1.1\%} & \textbf{73.8} & \textbf{72.9} & \textbf{1.2\%} & \textbf{74.6} & \textbf{73.5} & \textbf{1.5\%} \\
\textbf{Georgia} & \textbf{68.5} & \textbf{67.7} & \textbf{1.2\%} & \textbf{69.2} & \textbf{68.4} & \textbf{1.2\%} & \textbf{69.8} & \textbf{68.1} & \textbf{2.4\%} \\
\textbf{INCART} & \textbf{70.3} & \textbf{67.2} & \textbf{4.4\%} & \textbf{71.0} & \textbf{68.7} & \textbf{3.2\%} & \textbf{71.5} & \textbf{69.4} & \textbf{2.9\%} \\

\hline
\end{tabular}
\end{table*}

The results in the table illustrate the impact of including the target dataset in pre-training (w/) versus excluding it (w/o). Across all self-supervised learning (SSL) methods (BYOL, SimCLR, and MAE), pre-training on the target dataset consistently improves performance, though the extent of improvement varies. The largest performance gains are observed in INCART and CPSC, where BYOL pre-training leads to a 4.4\% and 3.6\% improvement, respectively. This suggests that datasets with higher variability or different recording conditions benefit significantly from inclusion in pre-training. In contrast, Chapman and PTB-XL, which are more homogeneous and well-structured, show minimal improvement (<1\%), indicating that models trained on diverse datasets already generalize well to these benchmarks. Additionally, the performance increase is more pronounced for BYOL and MAE compared to SimCLR, suggesting that feature consistency and reconstruction-based approaches are more effective for ECG pre-training. These findings highlight the importance of dataset selection in pre-training and suggest that incorporating the target dataset into the pre-training phase is particularly beneficial for datasets with higher inter-sample variability.

\subsection{Effect of Training Data Size on ECG Foundation Model Performance}

To analyze the impact of training data size on model performance, we gradually increase the amount of training data from 1\% to 100\% in controlled increments. At each stage, the model is trained on a subset of the full dataset and evaluated on a fixed test set. This experiment provides insights into how data volume influences model convergence, generalization, and downstream performance. Key evaluation criteria include performance saturation points, learning efficiency, and the trade-off between data volume and model performance.

\begin{figure*}[!t]  
    \centering
    \includegraphics[width=\linewidth]{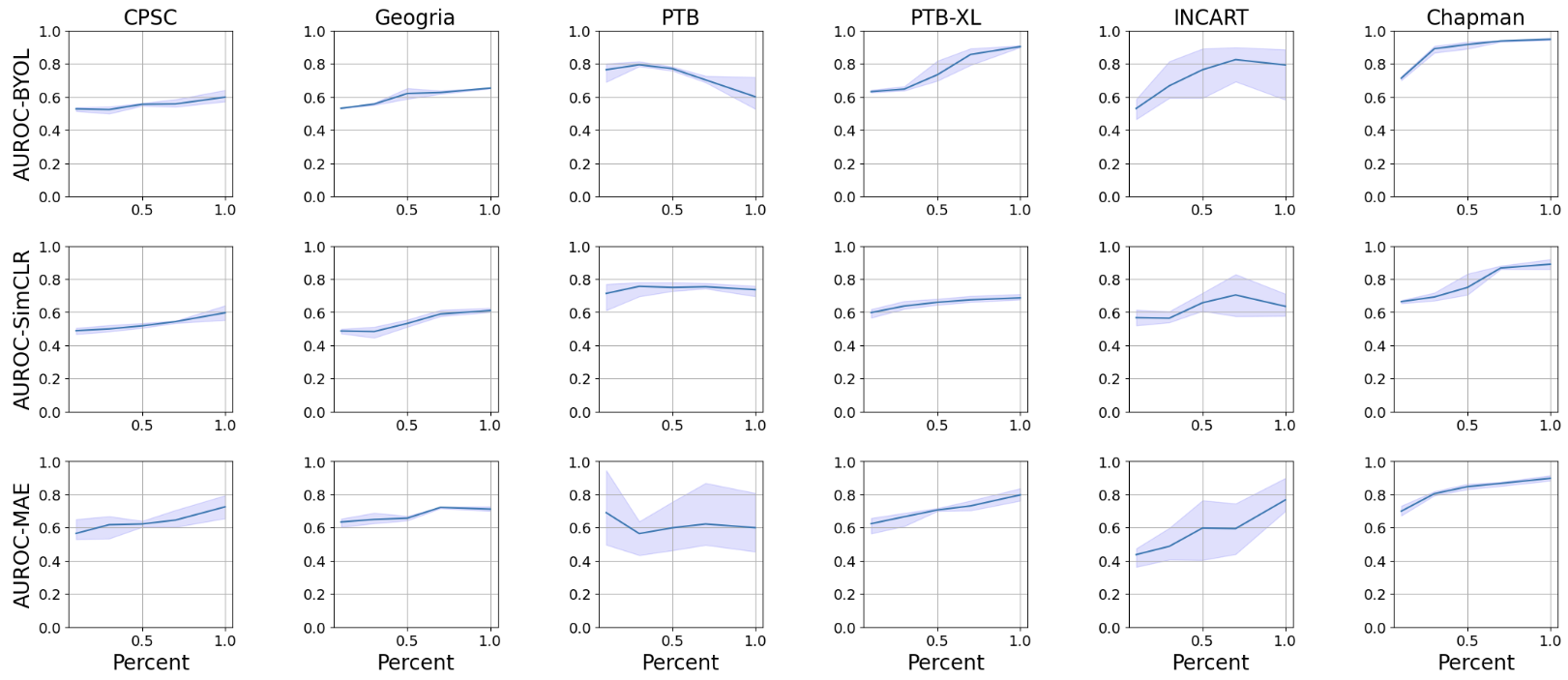}  
    \caption{Impact of Training Data Size on AUROC Performance Across Different Datasets and Self-Supervised Learning Methods}  
    \label{fig:example 2}  
\end{figure*}

The results of the data size scaling experiment reveal a clear relationship between training data volume and model performance. As the training data increases from 1\% to 100\%, all three methods exhibit consistent performance gains. Chapman and PTB-XL show a steady rise in AUROC, indicating that larger datasets significantly enhance model generalization. In contrast, PTB and Georgia exhibit fluctuations, suggesting potential overfitting or dataset-specific variations. CPSC and INCART datasets demonstrate lower AUROC values at smaller data sizes but benefit substantially from increased training data. The performance differences across SSL methods suggest that some approaches, such as MAE and BYOL, converge to high AUROC values more efficiently than SimCLR. Overall, these findings emphasize that while data scaling plays a crucial role in ECG model performance, dataset characteristics and SSL strategies also influence generalization, warranting further investigation into optimal pre-training data selection and augmentation techniques.

\section{Discussion}

Our study systematically explores the development of ECG foundation models (ECG-FMs) using a diverse set of publicly available datasets and evaluates the influence of various self-supervised learning (SSL) strategies on model generalization. The findings emphasize that large-scale, multi-center datasets significantly enhance the robustness of ECG-FMs, challenging the prevailing notion that high-performing AI models require proprietary data. Instead, when public data is curated, standardized, and leveraged effectively, it can yield clinically useful AI models capable of generalization across diverse patient populations and healthcare settings.

A key takeaway from our study is the impact of dataset diversity on model generalization. The iterative leave-one-dataset-out (LODO) experiment highlights that models trained on more heterogeneous datasets tend to generalize better, particularly for datasets with high variability, such as CPSC and INCART. This suggests that pre-training on diverse datasets is crucial for robustness, while models trained on homogeneous datasets may struggle with real-world variations. However, datasets like PTB-XL and Chapman, which are well-structured and consistent, exhibited minimal improvements when included in pre-training, indicating that foundation models trained on a broad dataset already perform well on them. These results suggest that the value of adding a dataset to pre-training depends on its variability and distinctiveness rather than just its size.

The comparison of different SSL strategies (BYOL, SimCLR, and MAE) provides insight into their strengths and limitations for ECG analysis. BYOL and MAE consistently outperform SimCLR, particularly in datasets with high variability, suggesting that feature consistency learning and generative reconstruction approaches are more suited for ECG representation learning than contrastive learning alone. This aligns with findings from natural language processing and computer vision, where contrastive learning requires large, high-quality datasets to be effective, whereas feature-consistency-based and generative models can learn meaningful representations with fewer constraints on data distribution. These insights could guide the selection of optimal pre-training strategies for future ECG-FM development, depending on the availability and characteristics of the training data.

The data scaling experiment further reveals that while performance improves as training data increases, performance saturation is reached around 60-70\% of the dataset size for BYOL and MAE, while SimCLR continues to improve beyond 80\%. This suggests that contrastive learning models rely more heavily on large datasets, whereas feature-consistency and generative models are more data-efficient. Importantly, overfitting or performance fluctuations in PTB and Georgia datasets at certain data sizes indicate that simply increasing dataset size is not always beneficial—data quality, diversity, and augmentation strategies play a crucial role in ensuring robust model performance. Future work should explore optimal dataset scaling strategies, including active learning approaches, selective sampling, and domain adaptation techniques to enhance ECG-FM efficiency.

Our study underscores the importance of standardization and benchmark development in ECG AI research. By integrating 1.2 million ECG recordings from multiple centers, we establish OpenECG, a large-scale benchmark that provides a standardized framework for evaluating ECG foundation models. This addresses a critical gap in the field, where prior ECG-FMs lacked systematic multi-center validation. Our benchmark serves as a publicly available reference point, enabling the research community to fairly compare models, refine SSL strategies, and drive future innovations in AI-driven cardiovascular diagnostics. The next steps involve extending OpenECG to incorporate multi-modal data, such as clinical notes, imaging, and genetic information, to develop a truly comprehensive cardiovascular AI model.

To sum up, this study provides empirical evidence for the feasibility of developing clinically relevant ECG foundation models using only public datasets. It highlights the need for dataset diversity, the effectiveness of different SSL methods, and the optimal strategies for dataset scaling. Our findings pave the way for scalable, generalizable, and accessible AI-driven ECG analysis, ultimately supporting equitable and widespread deployment of AI in cardiology.

\section{Conclusion}

This study introduces OpenECG, a comprehensive benchmark comprising 1.2 million 12-lead ECG recordings from diverse datasets, aimed at evaluating the effectiveness of SSL strategies for ECG foundation models (ECG-FMs). By aggregating data from multiple centers worldwide, we demonstrate that public datasets, when properly curated and standardized, can produce robust, clinically relevant ECG models without relying on proprietary data. 

Our results underscore the critical role of dataset diversity in enhancing the generalization capabilities of ECG-FMs. Models trained on heterogeneous datasets, such as those from CPSC and INCART, show superior performance across various cardiovascular disease classification tasks compared to those trained on homogeneous datasets. Among the SSL strategies explored, BYOL and MAE consistently outperform SimCLR, particularly in the presence of high dataset variability. These findings suggest that feature consistency and generative approaches are more effective for ECG representation learning than contrastive learning.

Additionally, we highlight the importance of dataset scaling and quality over mere volume. While increasing the training data size improves performance, saturation occurs at approximately 60-70\% of the total dataset for the BYOL and MAE models. This emphasizes the need for careful consideration of data characteristics and augmentation strategies to avoid overfitting and ensure model robustness.

By establishing OpenECG as a publicly available reference benchmark, we provide the research community with a standardized framework for evaluating and improving ECG-FMs. Looking ahead, expanding OpenECG to include multi-modal data, such as clinical notes and imaging, could further enhance the development of comprehensive cardiovascular AI models. Ultimately, this work contributes to the advancement of scalable, generalizable, and clinically meaningful AI-driven ECG analysis, paving the way for more equitable deployment of AI technologies in cardiology.

\section{REFERENCE}

\printbibliography[title={references}]
\end{document}